\documentclass[10pt,twocolumn,letterpaper]{article}

\usepackage{cvpr}              
\usepackage{graphicx}
\usepackage{tcolorbox}
\usepackage{framed}  
\usepackage{placeins}
\usepackage{afterpage} 
\usepackage{multirow}
\usepackage{verbatim}

\usepackage[normalem]{ulem}

\newcommand{\scope}{SCoPE}

\newcommand{\sdonefour}{Stable Diffusion 1-4}
\newcommand{\sdtwoone}{Stable Diffusion-2.1}

\usepackage{natbib}

\usepackage{booktabs}  
\usepackage{array}     
\usepackage{graphicx}  
\usepackage{adjustbox} 
\usepackage[]{tcolorbox}

\definecolor{cvprblue}{rgb}{0.21,0.49,0.74}
\usepackage[pagebackref,breaklinks,colorlinks,allcolors=cvprblue]{hyperref}

\def\paperID{*****} 
\def\confName{CVPR}
\def\confYear{2025}

\title{Progressive Prompt Detailing for Improved Alignment in Text-to-Image Generative Models}

\author{
Ketan Suhaas Saichandran$^{1}$\thanks{Equal contribution.}\quad\quad  Xavier Thomas$^{1}$\footnotemark[1]\quad\quad Prakhar Kaushik$^{2}$\quad\quad Deepti Ghadiyaram$^{13}$\thanks{Corresponding author.} \\
$^1$Boston University \quad
$^2$Johns Hopkins University \quad 
$^3$Runway\\
{\tt\small {\{ketanss, xthomas, dghadiya\}@bu.edu}\quad\quad pkaushi1@jhu.edu }}

\begin{document}
\maketitle


\begin{abstract}
Text-to-image generative models often struggle with long prompts detailing complex scenes, diverse objects with distinct visual characteristics and spatial relationships. In this work, we propose SCoPE (\textbf{S}cheduled interpolation of \textbf{Co}arse-to-fine \textbf{P}rompt \textbf{E}mbeddings), a training-free method to improve text-to-image alignment by progressively refining the input prompt in a coarse-to-fine-grained manner. Given a detailed input prompt, we first decompose it into multiple sub-prompts which evolve from describing broad scene layout to highly intricate details. During inference, we interpolate between these sub-prompts and thus progressively introduce finer-grained details into the generated image. Our training-free plug-and-play approach significantly enhances prompt alignment, achieves an average improvement of more than \textbf{+8} in Visual Question Answering (VQA) scores over the Stable Diffusion baselines on \textbf{83\%} of the prompts from the GenAI-Bench dataset.


\end{abstract}


\section{Introduction}
\label{sec:intro}
Text-to-image diffusion models~\cite{textimagediffusion} have made significant strides in generating high-quality generations from textual descriptions. Yet, they struggle to capture intricate details provided in long, detailed prompts describing complex scenes with multiple objects, attributes, and spatial relationships~\cite{paragraph_to_image, longclip}. When processing such prompts, these models often misrepresent spatial relations~\cite{derakhshani2023unlockingspatialcomprehensiontexttoimage,zhang2024compassenhancingspatialunderstanding}, omit crucial details~\cite{marioriyad2024attentionoverlapresponsibleentity}, or entangle distinct concepts~\cite{magnet_we_never_know, rahman2024visualconceptdrivenimagegeneration}. Several reasons contribute to this undesirable behavior. First, the text encoders used to condition the image generation process~\cite{openaiclip, t5} tend to compress a detailed textual description of varied lengths into a fixed-length representation, potentially leading to concept entanglement or information loss~\cite{clipproblem}. Second, biases in the pre-training data~\cite{laion} could be leading to favoring shorter prompts thereby degrading performance on long complex prompts~\cite{paragraph_to_image}.
\begin{figure}[!t]
\centering
\includegraphics[width=\linewidth]{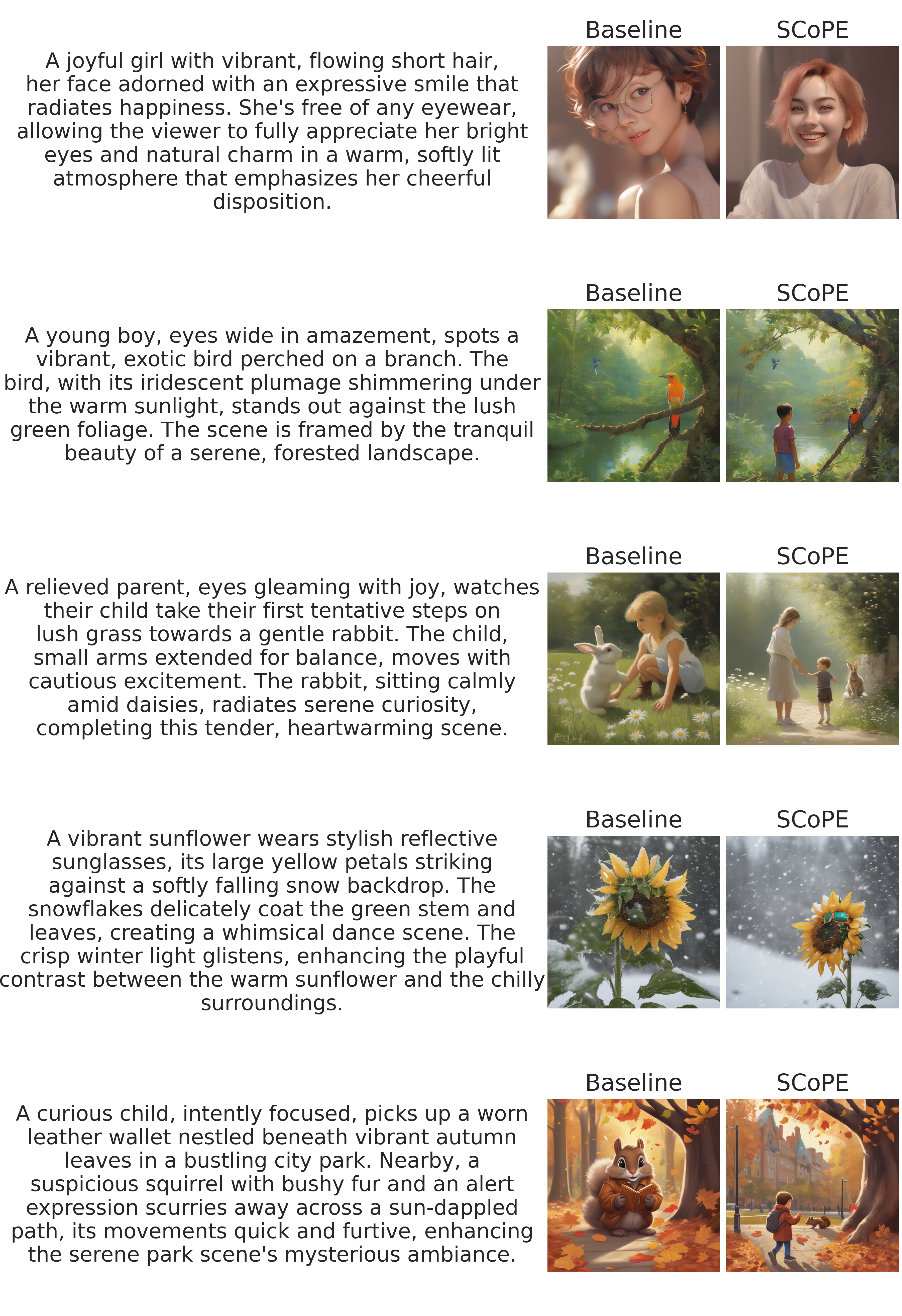}
\caption{\footnotesize \textbf{SCoPE (ours) vs {SDXL}~\cite{sdxl} for long, detailed prompts}. Note how SCoPE (right) captures details mentioned in the prompt better compared to {SDXL}. }
\label{fig:main}
\end{figure}

To address this limitation, ~\citet{longclip} extend the context length of the text encoder, allowing for better representation of longer prompts. While effective, this approach requires retraining on large-scale datasets, making it computationally expensive. Another line of work focuses on mitigating misalignment in the latent space~\cite{lcmis} by conditioning on individual concepts sequentially at different stages of the denoising process. However, this method primarily focuses on concept misalignment and entanglement in short prompts and does not focus on addressing the challenges with longer, detailed prompts. 
In this work, we propose \textbf{\scope} which stands for \textbf{S}cheduled interpolation of \textbf{Co}arse-to-fine \textbf{P}rompt \textbf{E}mbeddings. {\scope} is a training-free approach that improves alignment between the provided (long) prompt and the generated image in diffusion models. Our key idea is to dynamically break down the input prompt into a series of sub-prompts 
starting from a coarse-grained description that captures the global scene layout to more fine-grained details. 
We draw inspiration from the findings in~\citet{park2023understanding} that diffusion denoising is a progressive coarse-to-fine generation process, where initial timesteps establish low-frequency, global structures, while later steps introduce high-frequency, fine-grained details. Specifically, while all prior methods rely on a single static embedding of the entire input prompt, {\scope} interpolates between progressively detailed prompt embeddings throughout the denoising process, thus generating the global scene layout before gradually introducing finer-grained details. We extensively evaluate {\scope} against several open-sourced models and show that 
{\scope} improves prompt alignment for long complex prompts (obtained from the GenAI-bench dataset~\cite{genaibench}) (see Fig.~\ref{fig:main}) and achieves a \textbf{+8} improvement in VQA-based text-image alignment scores over Stable Diffusion~\cite{stablediffusion, sdxl} baselines. Notably, {\scope} is both training-free and easily extensible, requires minimal computational overhead of only $+0.7$ seconds per inference on $1$ A6000 GPU.

\section{Related work}


\noindent \textbf{Training on longer texts:}  Long-CLIP~\cite{longclip} expands the context length of CLIP-based text encoders to handle longer prompts, but requires explicit fine-tuning on long text descriptions. Similarly, \citet{paragraph_to_image} improve prompt alignment by fine-tuning both a Large Language Model (LLM) and the diffusion model, leveraging the LLM’s semantic comprehension capabilities to better encode the prompt and condition the generation process.
However, these approaches rely on high-quality dataset curation and require fine-tuning the diffusion model for prompt alignment. 
By contrast, SCoPE is training-free, efficient, and greatly improves prompt alignment.

\noindent \textbf{Interpolating text representations:} \citet{prompt_interpolation} explore interpolating between two prompt embeddings to control style and content in text-to-image diffusion models. While {\scope} explores a new direction, performing interpolation in a coarse-to-fine-grained manner, progressively refining text guidance throughout the generation process to improve alignment. 

\noindent \textbf{Addressing concept misalignment:} \citet{lcmis} highlights how text-to-image diffusion models struggle to accurately compose multiple distinct concepts, and often default to common co-occurring objects from training data. To mitigate this, they introduce concepts sequentially during generation. We build on this intuition and progressively add scene details throughout the denoising process, leading to better alignment with long, complex prompts.

\begin{figure}[!t]
\centering 
\includegraphics[width=0.5\textwidth]{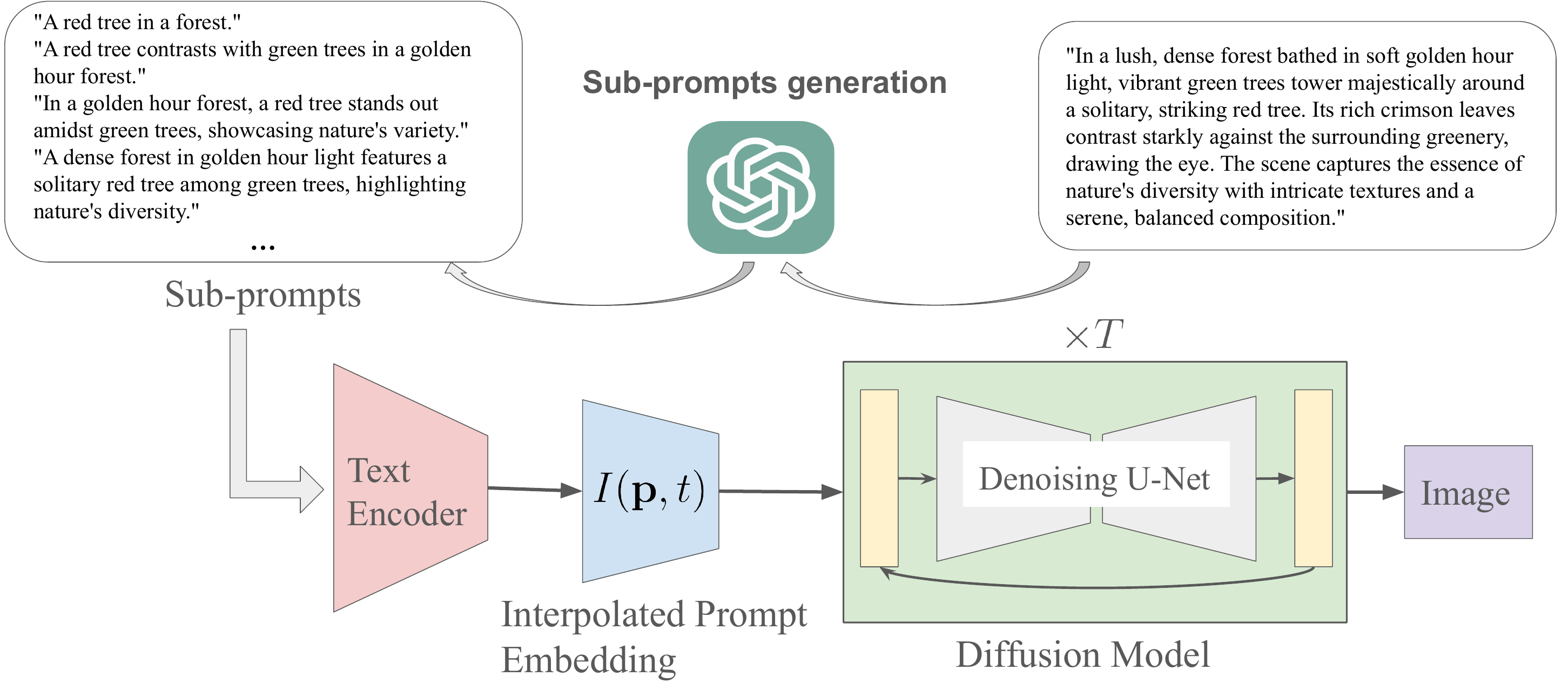}
\caption{ \footnotesize \textbf{Training-free approach of SCoPE}, where  we first decompose the input prompt into progressively detailed sub-prompts, then interpolates between their embeddings across timesteps, gradually introducing semantic details into the generations.} 
\label{fig:pipeline}
\end{figure}


\begin{table}[t]
    \centering
    \tiny 
    \renewcommand{\arraystretch}{1} 
    \setlength{\tabcolsep}{3pt} 
    \begin{tabular}{p{0.2cm} p{7.5cm}} 
        \toprule
        \textbf{Idx} & \textbf{Sub-Prompt} \\
        \midrule
        1 & A cat rides a skateboard on a city street. \\
        2 & A cat on a skateboard navigates a city street, balancing amidst people and cars. \\
        3 & An earless cat rides a skateboard in a busy street, casting shadows, maintaining balance against a city backdrop. \\
        4 & A sleek, earless cat skillfully rides a colorful skateboard on a bustling urban street, highlighted by sunlight shadows, showing agility and balance against a vibrant city backdrop. \\
        5 & A sleek, earless cat gracefully rides a colorful skateboard amidst a bustling urban street, effortlessly navigating between pedestrians and vehicles. Sunlight casts dramatic shadows, highlighting the feline's agile, streamlined form as it maintains perfect balance, capturing a sense of motion and boldness against the vibrant city backdrop. $\dagger$ \\
        \bottomrule
    \end{tabular}
    \caption{\footnotesize \textbf{Progressively detailed sub-prompts derived from the GenAI-Bench prompt \textit{A cat without visible ears is riding.}} $\dagger$ denotes the final prompt used to generate the baseline image. Refer to Fig.~\ref{fig:robots_yoga} for generation results.}
    \label{tab:robots_yoga_prompts}
\end{table}

\section{Approach}


We introduce {\scope} (depicted in Fig.~\ref{fig:pipeline}), a method for dynamically adjusting text conditioning in diffusion models. First, we describe how sub-prompts are generated from a given input text prompt, each representing a different level of scene granularity (Sec~\ref{sec:subprompt}). Next, we define an interpolation schedule to determine when each sub-prompt has the highest influence during denoising (Sec~\ref{sec:subprompt}). Finally, we describe our interpolation-based text conditioning approach, where the sub-prompts are blended over the denoising steps to guide the image generation process (Sec~\ref{sec:interpol}).

\subsection{Sub-prompt generation and interpolation schedule}
\label{sec:subprompt}

\noindent \textbf{Sub-prompt generation.}
We first use GPT-4o~\cite{gpt4o} to break down a given prompt into $n$ progressively detailed sub-prompts, each depicting the same scene with increasing level of detail (an example in Table~\ref{tab:robots_yoga_prompts}). We then obtain the CLIP embeddings~\cite{clip} of each sub-prompt such that $ \mathbf{p}_1$ corresponds to the embedding of the coarsest prompt and $\mathbf{p}_n$ to the final fine-grained prompt. 

\noindent \textbf{Interpolation schedule and interpolation period.} 
During image generation, {\scope} utilizes an interpolated representation (i.e., a weighted sum) of these sub-prompt embeddings for text-conditioning. To determine the timestep where each sub-prompt exerts its maximum influence during denoising, we define an \emph{interpolation schedule} that assigns each sub-prompt $\mathbf{p}_i$ to a specific timestep $q_i$ at which it has the highest influence on image generation. ($i \in \{1, 2, \dots, n\}$ represents the sub-prompt index). The schedule is initialized at $q_1$ set to 0, ensuring that the coarsest prompt $\mathbf{p}_1$ guides the early timesteps, where broad scene structures are formed, as noted in~\cite{semantics_vs_fidelity, park2023understanding}. We also define $q_n$ as the \emph{interpolation period} a hyperparameter that determines the timestep up to which interpolation is applied during denoising. We note that interpolation is applied only until timestep $q_n$, after which $\mathbf{p}_n$ serves as the sole text-conditioning input guiding the diffusion model.

\noindent\textbf{Constructing the interpolation schedule.} Instead of uniformly spacing the sub-prompts across the denoising timesteps, we adapt their placement based on the semantic similarity of their embeddings. Specifically, after selecting the hyperparameter $q_n$, we first set $q_1$ = 0. To determine the remaining timesteps $(q_2, \dots, q_{n-1})$, we calculate the Euclidean distance between consecutive embeddings, $d_i = \lVert \mathbf{p}_i - \mathbf{p}_{i-1} \rVert_2$, and ensure that the ratio $\frac{d_{i}}{q_{i} - q_{i-1}}$ remains constant $\forall i \in \{2, 3, \dots, n\}$. This ensures that semantically similar sub-prompts (i.e. those with smaller Euclidean distances) are assigned timesteps that are closer together, while sub-prompts with greater semantic differences are spaced further apart. We empirically find that this also facilitates a gradual refinement of details throughout the denoising process, which we define next. 

\subsection{Interpolation-based text conditioning}
\label{sec:interpol}



After defining the interpolation schedule, 
we use it to apply a Gaussian-based weighting mechanism at each denoising timestep $t \leq q_n$. Specifically, we define a Gaussian of standard deviation $\sigma$ centered at $q_i$. This aligns with our motivation where early timesteps benefit from broader, coarse guidance, while later timesteps favor sharper focus on fine-grained details. We define a weight to assign to each prompt embedding $\mathbf{p}_i$ at denoising timestep $t$ as $\alpha_{i,t} = \exp\left(-\frac{(t - q_i)^2}{2 \sigma^2}\right)$, where the denominator $\sigma$ controls the sharpness of the Gaussian function.
Following the symmetric decay of the Gaussian, during denoising, at each timestep $t \leq q_n$, weights assigned to earlier sub-prompts gradually decrease, while those for later sub-prompts increase. The weights $\alpha_{i,t}$ are then normalized to obtain $\alpha_{i,t}' = \frac{\alpha_{i,t}}{\sum_{j=1}^{n} \alpha_{j,t}}$. The final text embedding to condition at each timestep is computed as a weighted sum of sub-prompt embeddings, i.e., $I(\mathbf{p}, t) = \lVert \mathbf{p}_n\rVert \sum_{i=1}^{n} \alpha_{i,t}' {\mathbf{\hat{p}}}_i$, where $\hat{\mathbf{p}}_i = \frac{\mathbf{p}_i}{\lVert \mathbf{p}_i \rVert}$. The rescaling of magnitude by $\lVert \mathbf{p}_n\rVert$ is performed to ensure that the interpolated embedding lies on the hypersphere defined by the CLIP embedding space.







\begin{figure}[!t]
\centering
\includegraphics[width=0.45\textwidth]{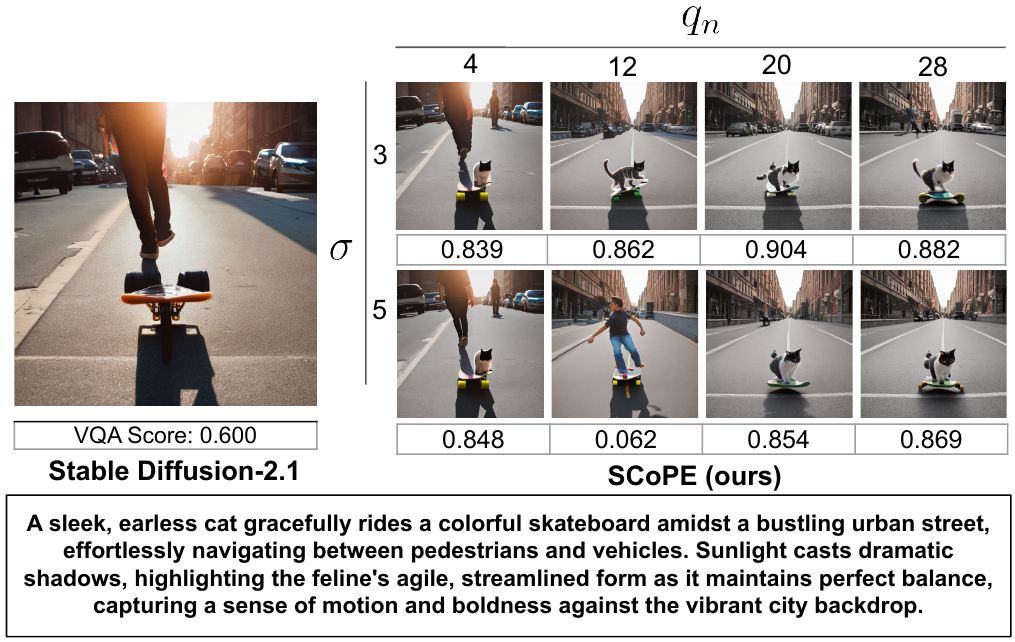}
\caption{\footnotesize \textbf{Effect of $\sigma$ and $q_n$ on image generation.} The figure illustrates how the variations in standard deviation ($\sigma$) and interpolation period ($q_n$) influence the generated images and their VQAScores. Smaller $q_n$ values (e.g., 4) preserve fine details (e.g., vehicles, pedestrians), while larger $q_n$ values (e.g., 20) emphasize broader structural elements, such as scene composition (e.g., cat, skateboard) and object interactions (e.g., riding). Similarly, smaller $\sigma$ values lead to images retaining more fine-grained details. See Sec.~\ref{sec:exp} for more details. The sub-prompts used for this example are provided in Table \ref{tab:robots_yoga_prompts}}.
\label{fig:robots_yoga}
\end{figure}

 
\noindent\textbf{Coarse-to-fine transition in {\scope}.} 
As denoising progresses, influence of coarser sub-prompts at later timesteps decreases. This results in a gradual shift in conditioning from coarse to fine details. The hyperparameter $\sigma$ further modulates this transition. Higher $\sigma$ values allows sub-prompts to maintain their influence longer, thereby yielding a more gradual transition. By contrast, lower values of $\sigma$ makes coarse sub-prompts to lose influence more quickly and resulting in a sharper transition to fine details. We note that for timesteps \( t > q_n \), \( I(\mathbf{p}, t) = \mathbf{p}_n \), i.e., interpolation is no longer applied, and the model conditions solely on the final fine-grained prompt embedding. This ensures that the image generation process is fully guided by the most detailed prompt, focusing on refining fine-grained details during the later denoising steps, during the fidelity-improvement phase, as discussed in~\citet{semantics_vs_fidelity}. Thus, the hyperparameter $q_n$ controls when fine-grained details begin to influence the denoising process, determining how early these details start to guide image generation. Fig.~\ref{fig:robots_yoga} shows how $\sigma$ and $q_n$ impact image generation.

\subsection{Evaluation Setup}
\noindent \textbf{Dataset:} We evaluate our approach using prompts derived from GenAI-Bench~\cite{genaibench}, which contains 1600 prompts with tags representing spatial relation, counting, negation, etc. These prompts have an average token length of $14.8$, and often lack fine-grained details. To ensure we test our model on longer, more detailed prompts, we applied the method from~\citet{prompt_interpolation} and adopt prompt enhancement, and to target a length of around $50$ words, to align within 75 tokens of the CLIP Text encoder. This step results in increasing the average token length to $69.3$. Specifically, we use GPT-4o~\cite{gpt4o} to generate a more detailed version of each prompt. Subsequently, we then use GPT-4o to decompose the enhanced prompt to return four variations, each capturing the same scene at a different level of detail. The prompts used to generate both the enhanced and simplified versions are provided in the appendix. We use the final fine-grained prompt to generate the baseline image generations and to obtain the evaluation scores described next.\\
\noindent \textbf{Metrics:} We evaluate the alignment between generated images and input text prompts (i.e. the fine-grained prompt) using VQAScore~\cite{vqascore} and CLIP-Score~\cite{clipscore} as our primary evaluation metrics. 
While CLIP-Score measures the cosine similarity between image and text embeddings, VQAScore~\cite{vqascore} uses a Visual Question Answering (VQA)~\cite{antol2015vqa} model to produce an alignment score by computing the probability of a ``Yes" answer to a simple ``Does this figure show \{input prompt\}?" question. Despite its simplicity, \citet{vqascore} demonstrates that VQAScore outperforms other methods in providing the most reliable text-image alignment scores, particularly for complex prompts. 
We also report ``Win\%" as the percentage of prompts where {\scope}-generated images outperform the baseline model.

\section{Experiments} 
\label{sec:exp}



\begin{table}[t]
\centering
\small 
\begin{tabular}{lcccc}
\toprule
 & \multicolumn{2}{c}{VQAScore~\cite{vqascore}} & \multicolumn{2}{c}{CLIP-Score~\cite{clipscore}} \\
 \hline
Model   & Mean & Win\% & Mean & Win \% \\
\midrule
SCoPE-v2-1 & \textbf{87.3} & 83.88\% & \textbf{34.9} & 77.56\% \\
SD-v2-1 & 79.2 & - & 33.6 & - \\
\hline
SCoPE-v1-4 & \textbf{84.7} & 83.44\% & \textbf{34.5} & 74.38\% \\
SD-v1-4 & 76.6 & - & 33.3 & - \\
\hline

SCoPE-XL & \textbf{87.7} & 73.00\% & \textbf{35.3} & 65.56\% \\
SDXL & 82.9 & - & 34.6 & - \\
\bottomrule
\end{tabular}
\caption{\footnotesize \textbf{Comparison of mean VQA Scores and CLIP Scores between {\scope} and baseline models.} Win\% indicates the percentage of prompts where {\scope}-generated images outperform the baseline. We observe that {\scope} consistently improves over the baselines, regardless of the model.}
\label{tab:comparison}
\end{table}

\begin{figure}[!t]
  \centering
  \includegraphics[width=1.0\linewidth]{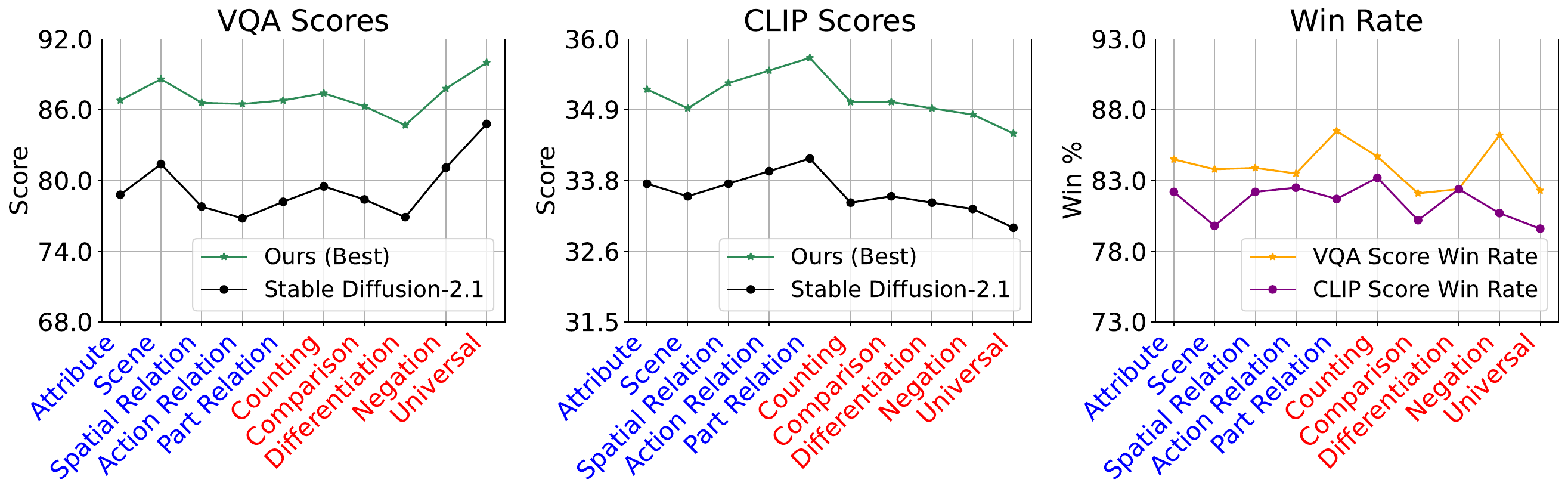}
\caption{ \footnotesize \textbf{Comparison between {\sdtwoone} and SCoPE across different prompt tags in GenAI-Bench~\cite{genaibench}.} The first five tags (Attribute, Scene, Action Relation, Part Relation, Counting Comparison) are categorized as ``Basic,” while the remaining tags (Differentiation, Negation, Universal) fall under the ``Advanced” category. We observe that {\scope} consistently outperforms the baseline {\sdtwoone} across both basic and advanced prompt categories.
}
\label{tab:tags}
\end{figure}

We evaluate {\scope} as a plug-and-play approach against {\sdonefour}~\cite{stablediffusion}, {\sdtwoone}~\cite{stablediffusion}, and SDXL~\cite{sdxl}. The total number of inference sampling steps were set to $50$ for all models. For each prompt, we generate 8 candidate output images using SCoPE with initial standard deviation ${\sigma}_{0}$ $\in$  \{3, 5\} and interpolation period $q_n$ $\in$ \{4, 12, 20, 28\}, as defined in Sec.~\ref{sec:interpol}. All experiments were carried out on $1$ A6000 GPU.

\noindent \textbf{Note on generating candidate outputs.} We conduct an empirical study across all 1600 prompts to examine whether VQA Scores~\cite{vqascore} correlate with the hyperparameters ($\sigma_0$, $q_n$) and found no clear pattern. In other words, given an input prompt, there is no clear way to predict in advance which setting will yield the most well-aligned generation. To account for this variability, we generate eight candidate outputs per prompt and evaluate them to select the most well-aligned result, described next.\\
\noindent \textbf{Quantitative results.} As shown in Table~\ref{tab:comparison}, {\scope} consistently improves text-image alignment, achieving higher VQA Scores and CLIP scores compared to the baselines. For {\sdtwoone}, {\scope} achieves a mean VQAScore of $\mathbf{87.3}$ across all 1600 prompts derived from GenAI-Bench, outperforming the baseline score of $79.2$. We also observe that {\scope} achieves an $\mathbf{83.88\%}$ win rate, indicating that in over $83\%$ of prompts, {\scope}-generated images were more aligned with the input text prompt (as measured by VQAScore). A similar trend is observed for CLIP Score, where {\scope} achieves a mean score of $\mathbf{34.9}$, surpassing the baseline score of $33.6$, with a $77.56\%$ win rate. For {\sdonefour} and SDXL, {\scope} increases VQAScore to $\mathbf{84.7}$ and $\mathbf{87.7}$, with win rates of $83.44\%$ and $73.00\%$, respectively. In Table~\ref{tab:tags} we show that {\scope} outperforms the baseline model on varied prompt categories, such as ``Spatial Relation," ``Counting," ``Negation," etc.\\

\section{Conclusion}

We propose {\scope}, a simple yet effective, training-free plug-and-play method that improves alignment in text-image generative models, particularly for long and detailed prompts. Our approach offers a lightweight solution that can be seamlessly integrated into existing pipelines. Future work may focus on reducing reliance on candidate outputs and extending applicability to broader generative tasks.

\newpage

{
    \small
    \bibliographystyle{ieeenat_fullname}
    \bibliography{main}
}

\clearpage
\setcounter{page}{1}
\maketitlesupplementary

\subsection*{Enhancement Prompt} 

\begin{tcolorbox}[colback=gray!5, colframe=black!70, sharp corners=southwest, fontupper=\ttfamily\small]
You are an expert at refining prompts for image generation models. Your task is to enhance the given prompt extensively by adding descriptive details and quality-improving elements, while maintaining the original intent and core concept.
Follow these guidelines:\\
1. Preserve the main subject and action of the original prompt.\\
2. Add specific, vivid details to enhance visual clarity.\\
3. Incorporate elements that improve overall image quality and aesthetics.\\
4. Keep the prompt concise and avoid unnecessary words.\\
5. Use modifiers that are appropriate for the subject matter.\\
6. The prompt should contain around 50 words.\\

Example modifiers (use as reference, adapt based on some aspect that's suitable for the original prompt):\\
- Lighting: "soft golden hour light", "dramatic chiaroscuro", "ethereal glow"\\
- Composition: "rule of thirds", "dynamic perspective", "symmetrical balance"\\
- Texture: "intricate details", "smooth gradients", "rich textures"\\
- Color: "vibrant color palette", "monochromatic scheme", "complementary colors"\\
- Atmosphere: "misty ambiance", "serene mood", "energetic atmosphere"\\
- Technical: "high resolution", "photorealistic", "sharp focus"\\

The enhanced prompt should be short, concise, direct, avoid unnecessary words and written as it was a human expert writing the prompt.\\
Output only one enhanced prompt without any additional text or explanations.
\end{tcolorbox}

\subsection*{Simplification Prompt}
\begin{tcolorbox}[colback=gray!5, colframe=black!70, sharp corners=southwest, fontupper=\ttfamily\small]
You are an expert at simplifying image descriptions. Your task is to simplify the description in 4 levels by removing any unnecessary words and phrases, while maintaining the original intent and core concept of the description.\\

Follow these guidelines:\\
1. Output a python list of 4 such prompts with increasing levels of simplification.\\
2. Preserve the main subject of the original description.\\
3. Remove all any unnecessary words and phrases.\\

Example:\\

Description:\\
A pristine sandy beach under a soft golden hour light, with fine grains of sand glistening in the warm sunlight. The shore is framed by gentle waves lapping at the dunes, creating a serene mood. Sparse details of seashells and driftwood add texture, inviting tranquility and relaxation.\\

Output:\\
\verb|[|'A beach with shiny sand and waves.', 'A beach at sunset with shiny sand. Waves hit dunes. Seashells and driftwood add texture.', 'A beach at golden hour with glistening sand. Waves lap at dunes, creating calm. Seashells and driftwood add texture.', 'A sandy beach at golden hour, with glistening sand. Gentle waves lap at the dunes, creating a serene mood. Seashells and driftwood add texture.'\verb|]|\\
\end{tcolorbox}

\clearpage
\subsection*{Tag-wise Scores}
\begin{table}[!h]
\centering
\begin{tabular}{|c|c|c|c|c|c|c|c|}
\hline
\multirow{3}{*}{Level} & \multirow{3}{*}{Tag}   & \multicolumn{2}{c|}{VQA Score} & \multicolumn{2}{c|}{CLIP Score} & \multicolumn{2}{c|}{Win \%} \\ \cline{3-8}
         &                   & SCoPE  & SD-v2-1  & SCoPE  & SD-v2-1 & VQA & CLIP \\ \cline{1-8}
\multirow{5}{*}{Basic}    
         & Attribute         & 0.8683 & 0.7878 & 0.3521 & 0.3367 & 84.53\% & 82.22\% \\
         & Scene             & 0.8857 & 0.8145 & 0.3490 & 0.3350 & 83.78\% & 79.80\% \\
         & Spatial Relation  & 0.8661 & 0.7783 & 0.3535 & 0.3372 & 83.87\% & 82.19\% \\
         & Action Relation   & 0.8649 & 0.7675 & 0.3549 & 0.3387 & 83.51\% & 82.46\% \\
         & Part Relation     & 0.8678 & 0.7820 & 0.3572 & 0.3412 & 86.46\% & 81.66\% \\ \hline
\multirow{5}{*}{Advanced} 
         & Counting          & 0.8743 & 0.7949 & 0.3501 & 0.3340 & 84.66\% & 83.19\% \\
         & Comparison        & 0.8632 & 0.7845 & 0.3505 & 0.3351 & 82.10\% & 80.25\% \\
         & Differentiation   & 0.8472 & 0.7687 & 0.3489 & 0.3342 & 82.43\% & 82.43\% \\
         & Negation          & 0.8780 & 0.8105 & 0.3483 & 0.3326 & 86.17\% & 80.69\% \\
         & Universal         & 0.8997 & 0.8478 & 0.3451 & 0.3302 & 82.31\% & 79.59\% \\ \hline
\end{tabular}
\caption{VQA and CLIP Scores on different tags from the GenAI-Bench~\cite{genaibench} dataset for {\scope} and {\sdtwoone} (SD-v2-1). Absolute improvements are consistent across all categories.}
\label{tab:performance_metrics_comparison}
\end{table}

\subsection*{More Qualitative Results}
\begin{figure*}[!t]
\centering
\includegraphics[width=\textwidth]{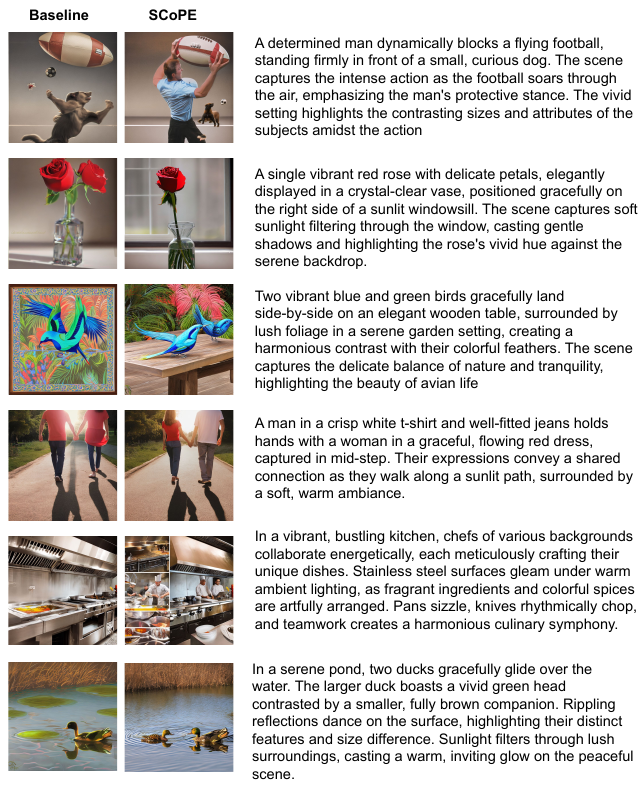}
\caption{More examples to compare {\scope} generated images and the images generated from the baseline {\sdtwoone}.}
\end{figure*}

\end{document}